\documentclass{article}

\usepackage{microtype}
\usepackage{xcolor}
\usepackage{booktabs}
\usepackage{multirow}
\usepackage{graphicx}
\usepackage{float}
\usepackage{subcaption}
\usepackage{booktabs} % for professional tables
\usepackage{amsmath}
\usepackage{hyperref}

\usepackage[preprint]{icml2026}

\usepackage{amsmath}
\usepackage{amssymb}
\usepackage{mathtools}
\usepackage{amsthm}

\usepackage[capitalize,noabbrev]{cleveref}

\theoremstyle{plain}

\theoremstyle{definition}

\theoremstyle{remark}

\usepackage[textsize=tiny]{todonotes}

\icmltitlerunning{VLGOR: Visual-Language Knowledge Guided Offline Reinforcement Learning for Generalizable Agents}

\begin{document}

\onecolumn
  \icmltitle{VLGOR: Visual-Language Knowledge Guided Offline Reinforcement \\Learning for Generalizable Agents}

\icmlsetsymbol{equal}{*}

\begin{icmlauthorlist}
  \icmlauthor{Pengsen Liu}{equal,nju}
  \icmlauthor{Maosen Zeng}{equal,cic}
  \icmlauthor{Nan Tang}{nju}
  \icmlauthor{Kaiyuan Li}{nju}
  \icmlauthor{Jing-Cheng Pang}{nju}
  \icmlauthor{Yunan Liu}{cic}
  \icmlauthor{Yang Yu}{nju}
  \icmlcorrespondingauthor{Yang Yu}{yuy@nju.edu.cn}
\end{icmlauthorlist}

\icmlaffiliation{nju}{National Key Laboratory for Novel Software Technology, Nanjing University, China \& School of Artificial Intelligence, Nanjing University, China}
\icmlaffiliation{cic}{Computational Intelligence Center (CIC), School of Computer and Artificial Intelligence, Shandong Jianzhu University, Jinan, 250101, China}

\printAffiliationsAndNotice{\icmlEqualContribution}
\vspace{3em}
\begin{abstract}
Combining Large Language Models (LLMs) with Reinforcement Learning (RL) enables agents to interpret language instructions more effectively for task execution. However, LLMs typically lack direct perception of the physical environment, which limits their understanding of environmental dynamics and their ability to generalize to unseen tasks. To address this limitation, we propose Visual-Language Knowledge-Guided Offline Reinforcement Learning (VLGOR), a framework that integrates visual and language knowledge to generate imaginary rollouts, thereby enriching the interaction data. The core premise of VLGOR is to fine-tune a vision-language model to predict future states and actions conditioned on an initial visual observation and high-level instructions, ensuring that the generated rollouts remain temporally coherent and spatially plausible. Furthermore, we employ counterfactual prompts to produce more diverse rollouts for offline RL training, enabling the agent to acquire knowledge that facilitates following language instructions while grounding in environments based on visual cues. Experiments on robotic manipulation benchmarks demonstrate that VLGOR significantly improves performance on unseen tasks requiring novel optimal policies, achieving a success rate over 24\% higher than the baseline methods.
\end{abstract}

\section{Introduction}

Developing agents capable of performing diverse manipulation tasks is a key hallmark of machine intelligence. In recent years, Reinforcement Learning (RL)~\cite{mnih2015human,vinyals2019grandmaster,li2024discovering} has emerged as an effective paradigm for training autonomous agents in interactive environments. Despite significant progress, RL remains largely constrained by the diversity and coverage of available interaction data. As a result, agents often struggle with tasks that are not represented in the training data and exhibit limited generalization to novel tasks or minor task variations~\cite{hafner2025mastering}. Therefore, building generalist agents that can respond to diverse user instructions~\cite{zheng2025imanip} and generalize to unseen tasks~\cite{tian2023decompose} remains a challenging problem.

Offline RL~\cite{fujimoto2021minimalist,ran2023policy,wang2023offline,ijcaiLiLCL24} addresses generalization by training agents over a distribution of offline tasks to improve performance on new tasks. However, many existing approaches rely on high-quality samples~\cite{yang2023essential} or warm-up exploration~\cite{gao2023context} to infer task beliefs at test time, while such supervision signals are costly and often infeasible to obtain in advance for unseen tasks~\cite{wang2024meta}. This limitation highlights the need for a supervision source that is cost-effective, broadly available, and highly scalable. Natural language~\cite{mihalcea2024developments}, as one of the richest and most accessible carriers of human knowledge, provides a promising alternative. Recent advances in Large Language Models (LLMs) enable the acquisition of general knowledge from large-scale text corpora, supporting generalization at the semantic level. In contrast, offline RL learns primarily from environment interactions, resulting in decision-level representations~\cite{brohan2023can,liu2024exploring} that are tightly coupled to specific dynamics and largely independent of language. Motivated by this gap, recent studies explore integrating language into decision making, including using LLMs as policies~\cite{szot2023large,li2024driving} and as sources of reward supervision~\cite{yang2024text2reward,citation0}.

\begin{figure*}[h!]
  \begin{center}
    \centerline{\includegraphics[width=0.85\linewidth]{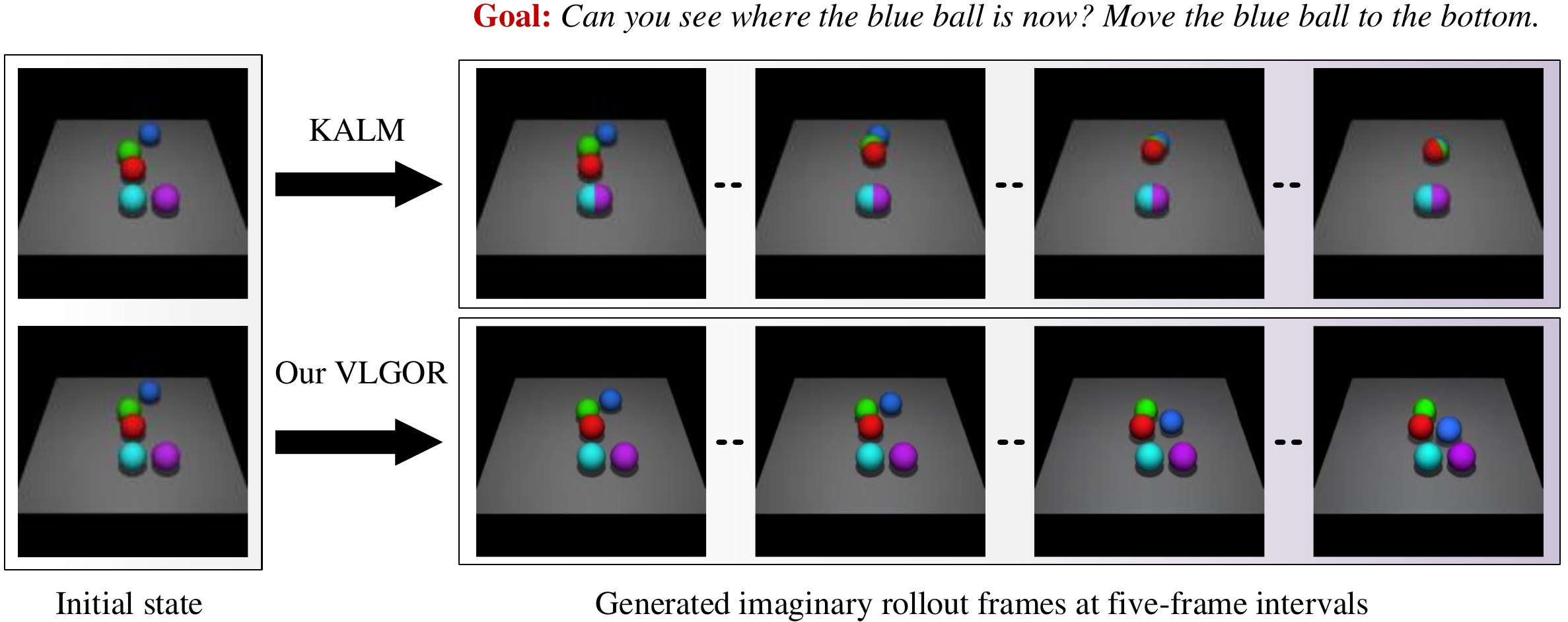}}
    \caption{
      An example comparison of imaginary rollouts generated by KALM~\cite{pang2024kalm} and our VLGOR. Relying solely on language knowledge, KALM does not explicitly model the spatial states of multiple objects. In contrast, VLGOR generates more realistic rollouts with smoother and physically plausible object motions. 
    }
    \label{fig1}
  \end{center}
\end{figure*}

Despite the rich semantic knowledge provided by LLMs, relying solely on textual representations for low-level action generation or the synthesis of imaginary rollouts remains fundamentally limited. LLMs are designed to operate on discrete language tokens, whereas physical environments evolve according to continuous numerical dynamics, such as positions, orientations, and velocities. Most existing approaches~\cite{carta2023grounding,tan2024true,pang2024kalm,glossop2025cast} attempt to map abstract language tokens into continuous vector spaces; however, without direct perceptual grounding in the physical environment, these representations struggle to faithfully capture the underlying temporal dynamics and spatial structure. As a result, the generated trajectories may appear semantically plausible but often violate physical constraints (e.g., object penetration or unrealistic motion), making them unsuitable for precision-critical low-level control. These limitations indicate that language-only modeling provides insufficient environmental grounding for offline reinforcement learning, thereby motivating the integration of visual perception with language understanding and naturally leading to visual-language models as a more physically consistent representation.

To address the aforementioned limitations, we leverage vision–language models to generate frame-level observations along each rollout. As shown in \cref{fig1}, we provide a representative example comparing trajectories produced by our method and KALM\footnote{KALM uses language to describe states for generating imaginary rollouts; for comparison, we render each frame using the MuJoCo physics engine~\cite{todorov2012mujoco}.}~\cite{pang2024kalm}. Compared with KALM, our method explicitly models physical constraints, resulting in more realistic trajectories. To generate diverse rollouts, we leverage counterfactual prompts as instructions to augment interaction data for offline reinforcement learning. This enables the agent to follow high-level instructions while respecting environmental dynamics, thereby improving its generalization to unseen tasks.

Our contributions are threefold. First, we propose Visual-Language Knowledge Guided Offline Reinforcement Learning (VLGOR), a unified framework that integrates vision–language models into offline RL, enabling agents to achieve precise low-level control through augmented imaginary rollouts. Second, we fine-tune the vision–language model and leverage counterfactual prompts as instructions, allowing the generated rollouts to exhibit greater diversity while remaining consistent with real-world physical constraints. Finally, we evaluate VLGOR on two robotic manipulation benchmarks, where the results demonstrate strong generalization to unseen tasks, achieving a 41.4\% success rate on 3,200 novel CLEVR-Robot goals.

\begin{figure*}[h!]
  \begin{center}
    \centerline{\includegraphics[width=0.85\linewidth]{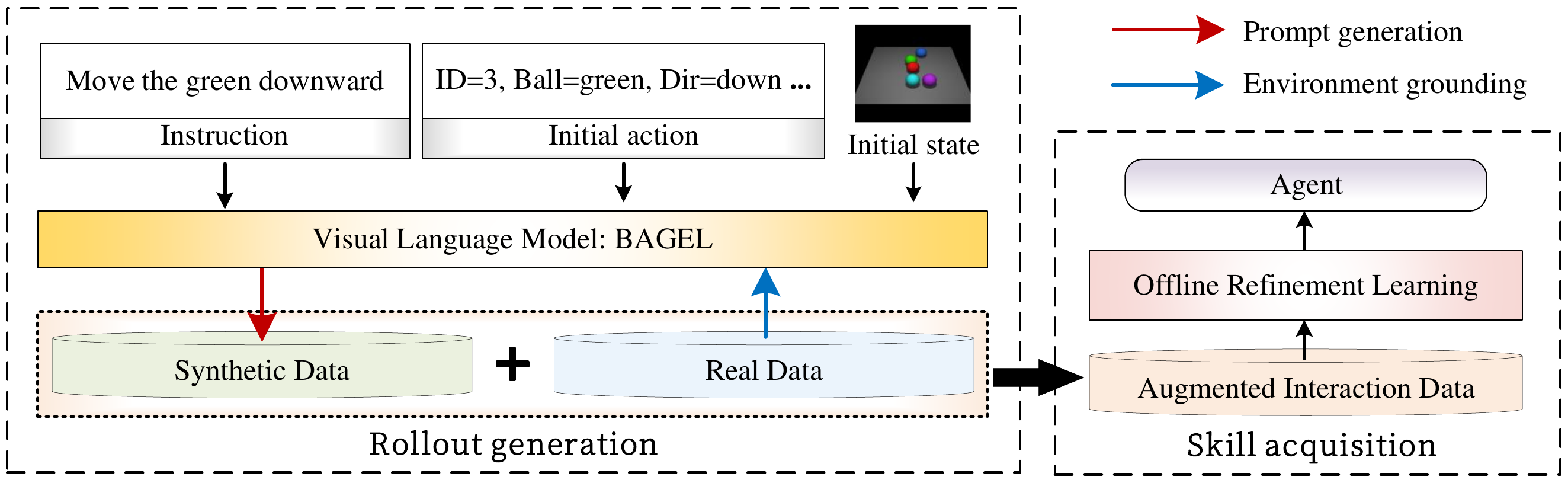}}
    \caption{
      Overall framework of the proposed VLGOR method. It consists of two stages: rollout generation for synthesizing imaginary rollouts from novel instructions, and skill acquisition for learning policies via offline reinforcement learning. 
    }
    \label{fig2}
  \end{center}
\end{figure*}

\section{Related Work}

\subsection{Offline Reinforcement Learning (RL)} Offline RL trains agents solely on pre-collected datasets, without requiring online interaction with the environment. Early studies extended behavior cloning by incorporating techniques such as sampling~\cite{hong2023beyond} and representation learning~\cite{ishfaq2024offline}, thereby advancing the development of Offline RL. Nevertheless, a fundamental challenge remains: how to derive effective policies from datasets that are limited in size or suffer from distributional shift. COMBO~\cite{yu2021combo} combines offline datasets with model-generated rollouts to train value functions, while ReDM~\cite{jia2024policy} simulates diverse environment rollouts by training multiple environment models. More recently, contextual reinforcement learning methods~\cite{huang2024context,wu2025mixture} leverage the in-context learning capability of transformers to improve generalization in RL. Despite their empirical success, these approaches typically rely on high-quality samples or domain knowledge to infer task beliefs; when specific experiences are absent from the dataset, agents may fail to perform adequately in such unseen scenarios. Motivated by this limitation, we explore the use of imaginary rollouts generated by vision–language models as a broader and complementary source of supervision.

\subsection{Large Language Models (LLMs) for RL}
LLMs have shown strong capabilities in natural language understanding and processing~\cite{singhal2023large,chang2024survey}, motivating a promising research direction: how to leverage LLMs to improve RL in interactive tasks.  Existing approaches mainly follow two technical lines. One line uses LLMs to automatically generate reward functions~\cite{yang2024text2reward,citation0}, providing denser and more informative supervision for RL training. The other line directly uses LLMs as policies and trains them with RL, e.g., TWOSOME~\cite{tan2024true} in text-based games and LLaRP~\cite{szot2023large} for embodied tasks. To improve language–environment alignment, recent work has explored stronger mechanisms for learning language–action associations. CAST~\cite{glossop2025cast} produces multiple language instructions and corresponding action labels for the same observations, while KALM~\cite{pang2024kalm} uses supervised fine-tuning to translate between language-described skills and rollout data. In this work, we propose a new approach that leverages vision–language models to generate imaginary rollouts, enabling the training of knowledge-driven, environment-aware decision agents.

\section{Method}
In this section, we introduce the proposed VLGOR method, whose framework is illustrated in Figure 2. We first provide a formal definition of the problem, followed by a detailed description of the two core steps of VLGOR: (1) \textbf{rollout generation}, which produces imaginary rollouts based on novel instructions to augment the interaction data; and (2) \textbf{skill acquisition}, which trains the policy via offline reinforcement learning to enable the agent to acquire the skills necessary to accomplish various instructions.

\subsection{Problem Formulation}
Let $\mathcal{D}$ denote a dataset collected from the interaction environment, primarily used for offline RL. It contains a set of natural language instructions (goals), denoted as $G$, along with the corresponding rollouts that achieve them, denoted as $R$. Suppose $\mathcal{D}$ contains $K$ instructions; then, we have $\mathcal{D}{=}{(G_k, R_k)}_{k=1}^{K}$, where $R_k{=}(s_0^k, a_0^k, s_1^k, a_1^k, \dots)$, and $(s_i^k, a_i^k)$ represents the state and action at the $i$-th timestep of the rollout. In offline RL, each rollout captures how an agent gradually achieves its corresponding goal while maximizing cumulative reward.

The core of VLGOR is to leverage both visual and language knowledge to construct an augmented dataset $\mathcal{D}_a$, which extends beyond the original interaction data $\mathcal{D}$ by incorporating diverse rollouts. This enriched dataset enables the agent to improve its decision-making generalization on unseen tasks. As illustrated in \cref{fig2}, VLGOR consists of two key components: rollout generation and skill acquisition. In the rollout generation stage, a vision-language model is fine-tuned to align with the environment and produce diverse rollouts conditioned on instructions and initial states. In the skill acquisition stage, both the original and augmented rollouts, along with their instructions, are used in offline RL to train an agent capable of acquiring new skills. Further details on these two stages are provided in \cref{sec32} and \cref{sec33}, respectively.

\begin{figure*}[t]
  \begin{center}
    \centerline{\includegraphics[width=0.9\linewidth]{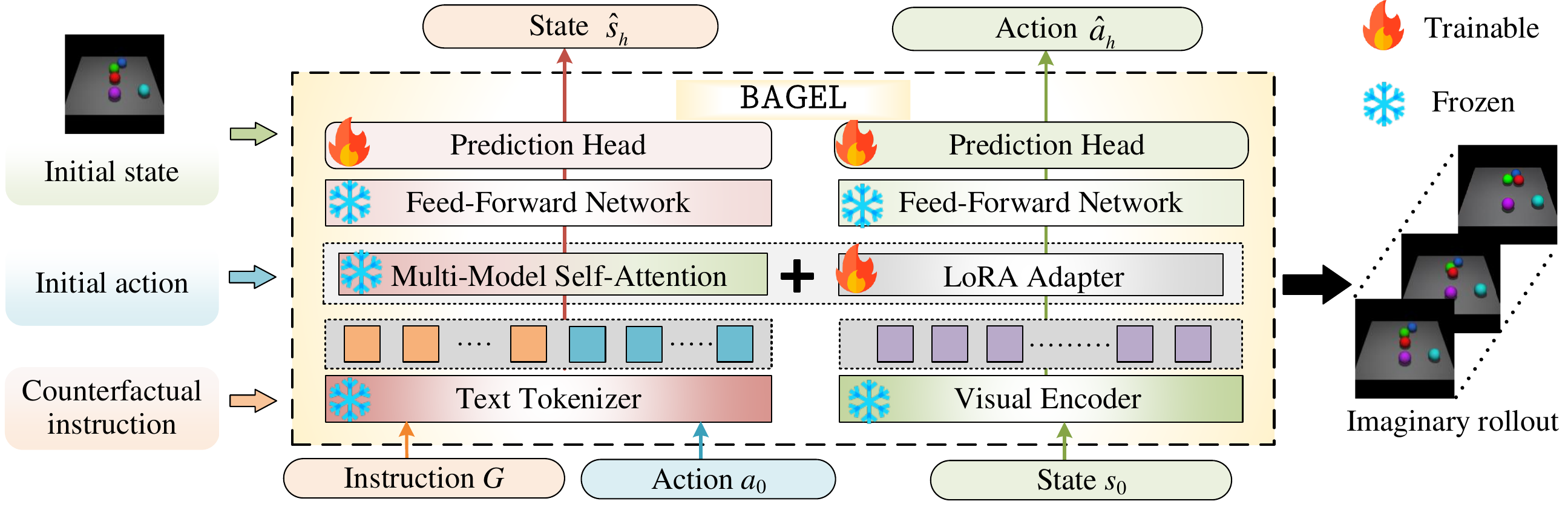}}
    \caption{
     Illustration of BAGEL fine-tuning optimization via vertical data flow and rollout generation via horizontal data flow. 
    }
    \label{fig3}
  \end{center}
\end{figure*}

\subsection{Rollout Generation via Counterfactual Prompts}\label{sec32}

In this study, we employ the vision–language model BAGEL~\cite{deng2025emerging} to augment interaction data. First, we fine-tune BAGEL for environment grounding. The model is optimized to (1) interpret the semantic content of textual instructions and (2) perceive physical and dynamic constraints from visual inputs. This grounding enables VLGOR to generate physically consistent rollouts conditioned on instructions and initial states. After fine-tuning, we leverage counterfactual prompts to generate diverse rollouts.

\cref{fig3} illustrates the fine-tuning and rollout generation process of BAGEL, which is built upon a mixture-of-transformers architecture. Textual and visual inputs are first encoded by a tokenizer and an encoder, respectively. The modality-specific experts then operate on a shared token sequence through a unified Multi-Modal Self-Attention (MMSA) module. Subsequently, the Feed-Forward Network (FFN) is duplicated into two parallel branches for textual and visual token prediction. During fine-tuning, trainable LoRA adapters~\cite{iclrHuSWALWWC22} are inserted into the MMSA module to facilitate effective cross-modal modeling. In addition, the parameters of the modality-specific prediction heads in both branches are optimized to reflect the independence of predictions across different modalities.

\textbf{VLM Fine-tuning for Environment Grounding.} We perform supervised fine-tuning on the dataset $\mathcal{D}$. For a given trajectory $R_k$, we extract a sub-trajectory consisting of the first $h$ steps, $(s_0^k, a_0^k, \dots, s_{h-1}^k, a_{h-1}^k)$, as a training sample. The instruction $G_k$ together with the action $a_0^k$ form the textual input $x_{\text{text}}^k$, while the image corresponding to the state $s_0^k$ serves as the visual input $x_{\text{vis}}^k$. The model then predicts the future state $\hat{s}_h^k$ and action $\hat{a}_h^k$ at step $h$ as follows:
\begin{equation}
(\hat{s}_h^k, \hat{a}_h^k) = f_\theta(x_{\text{vis}}^k, x_{\text{text}}^k),
\end{equation}
where $f_\theta$ denotes the BAGEL model, and $\theta$ represents the trainable parameters of the LoRA adapters and the prediction heads. Both the predicted visual states $\hat{s}_k^h$ and actions $\hat{a}_k^h$ are supervised using a Mean Squared Error (MSE) loss.

\textbf{Counterfactual Prompt-based Rollout Generation.}  
After fine-tuning, we generate diverse rollouts by leveraging counterfactual prompts. Each counterfactual instruction $G'_k$ consists of three components: a background description template $t$, a manipulation target $m$, and a behavior pattern $b$, i.e., $G'_k{=}\langle t,m,b\rangle$. We construct two types of counterfactual prompts by modifying $m$ and $b$:
\begin{itemize}
    \item \textit{Counterfactual target}: the behavior pattern $b$ is kept unchanged, while the manipulation target $m$ is replaced.
    \item \textit{Counterfactual behavior}: the manipulation target $m$ is kept unchanged, while the behavior pattern $b$ is altered.
\end{itemize}
Given these counterfactual instructions $G'_k$, the fine-tuned BAGEL model predicts the future state $\hat{s}_h^k$ and action $\hat{a}_h^k$, conditioned on the initial state $s_0^k$ and textual input $(G'_k, {a'}_0^k)$. This procedure generates augmented rollouts $R'_k$, thereby enriching the interaction data and improving the agent's ability to generalize to unseen tasks by adapting to variations in manipulation targets and behavior patterns.

\begin{figure*}[ht]
  \vskip 0.2in
  \begin{center}
    \centerline{\includegraphics[width=0.88\linewidth]{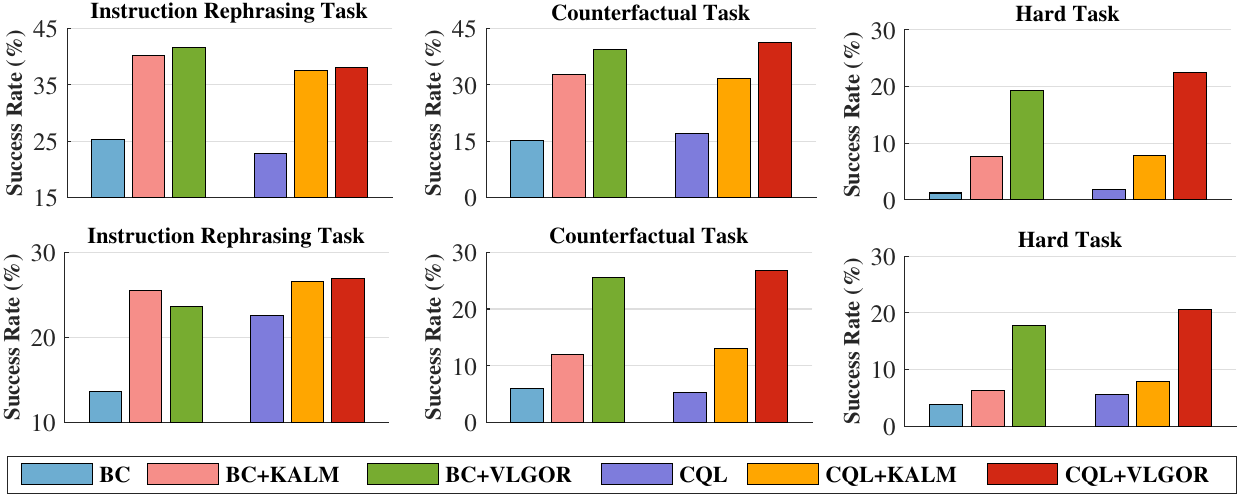}}
    \caption{
      Success rates (\%) of different methods on unseen tasks with varying difficulty levels.
The top row reports success rate on the CLEVR-Robot environment, while the bottom row reports success rate on the Meta-World environment. Results are averaged over the last five checkpoints, with error bars indicating half the standard deviation across three random seeds. 
    }
    \label{fig4}
  \end{center}
\end{figure*}

\subsection{Skill Acquisition through Offline RL}\label{sec33}

By leveraging counterfactual prompts, we generate augmented rollouts that extend beyond the original interaction data. VLGOR combines these generated rollouts with real trajectories to construct a richer dataset $D_t$ for offline RL. The objective is to learn a goal-conditioned policy $\pi(a \mid s, G)$ that enables the agent to acquire diverse skills conditioned on natural language instructions.

In the policy network, lightweight convolutional blocks encode the visual observation $s$, while BERT~\cite{devlin2019bert} encodes the natural language goal $G$. The resulting visual and textual feature representations are fused and jointly provided as input to the policy network. For policy optimization, VLGOR is, in principle, compatible with any offline RL algorithm and is trained on the combined dataset $D_t$. We consider two representative offline RL baselines~\cite{mediratta2023generalization}: Behavior Cloning (BC) and Conservative Q-Learning (CQL). BC performs direct imitation of actions in the dataset, whereas CQL learns a value-based policy to remain robust against out-of-distribution actions.

\textbf{Behavior Cloning (BC)} serves as a pure supervised baseline. Given the dataset $D_t$, the policy directly imitates the observed actions. For a discrete action space $\mathcal{A}$, the policy outputs logits, which are transformed into probabilities via softmax. The BC objective is:
\begin{equation}
\mathcal{L}_{\text{BC}}(\theta') = 
-\mathbb{E}_{(s, a, G) \sim D_t} 
\Big[ 
    \log \pi_{\theta'}(a \mid s, G) 
\Big],
\end{equation}
where $\theta'$ denotes the parameters of the policy network. After optimization, the learned policy $\pi_\theta(a \mid s, G)$ captures the conditional action distribution and provides a baseline for evaluating the benefits of VLM-generated rollouts in learning goal-conditioned actions.

\textbf{Conservative Q-Learning (CQL)} is a value-based offline RL algorithm designed to mitigate overestimation of action values for out-of-distribution actions. In a goal-conditioned setting, the Q-function is $Q_\phi(s,a,G)$, where $s$ is the visual state, $a$ is a discrete action, and $G$ denotes a natural language goal. CQL minimizes the standard Bellman regression loss augmented with a conservative regularization term:
\begin{equation}
\begin{split}
\mathcal{L}_{\text{CQL}}(\phi) = \alpha \Bigg(
& \mathbb{E}_{(s,G) \sim D_t} \Big[ 
\log \sum_{a} \exp Q_\phi(s,a,G) 
\Big] \\
& - \mathbb{E}_{(s,a,G) \sim D_t} \big[ Q_\phi(s,a,G) \big]
\Bigg),
\end{split}
\end{equation}
where $\alpha$ controls the strength of conservatism. The resulting Q-function is used to derive a goal-conditioned policy $\pi_\theta(a\mid s,G)$, providing a robust baseline for evaluating the benefit of VLM-generated rollouts.

\section{Experiment}
In this section, we conduct experiments to evaluate the effectiveness of the VLGOR method. The experiments are designed to answer the following questions:
\begin{itemize}
    \item How does VLGOR perform in offline RL compared with existing baseline methods (\cref{sec42})?
    \item How is the quality of rollouts generated by VLGOR through fine-tuning and prompting (\cref{sec43})?
    \item How do each component of VLGOR affect the overall performance of RL offline (\cref{sec44})?
\end{itemize}

\begin{figure*}[t]
  \vskip 0.2in
  \begin{center}  \centerline{\includegraphics[width=0.86\linewidth]{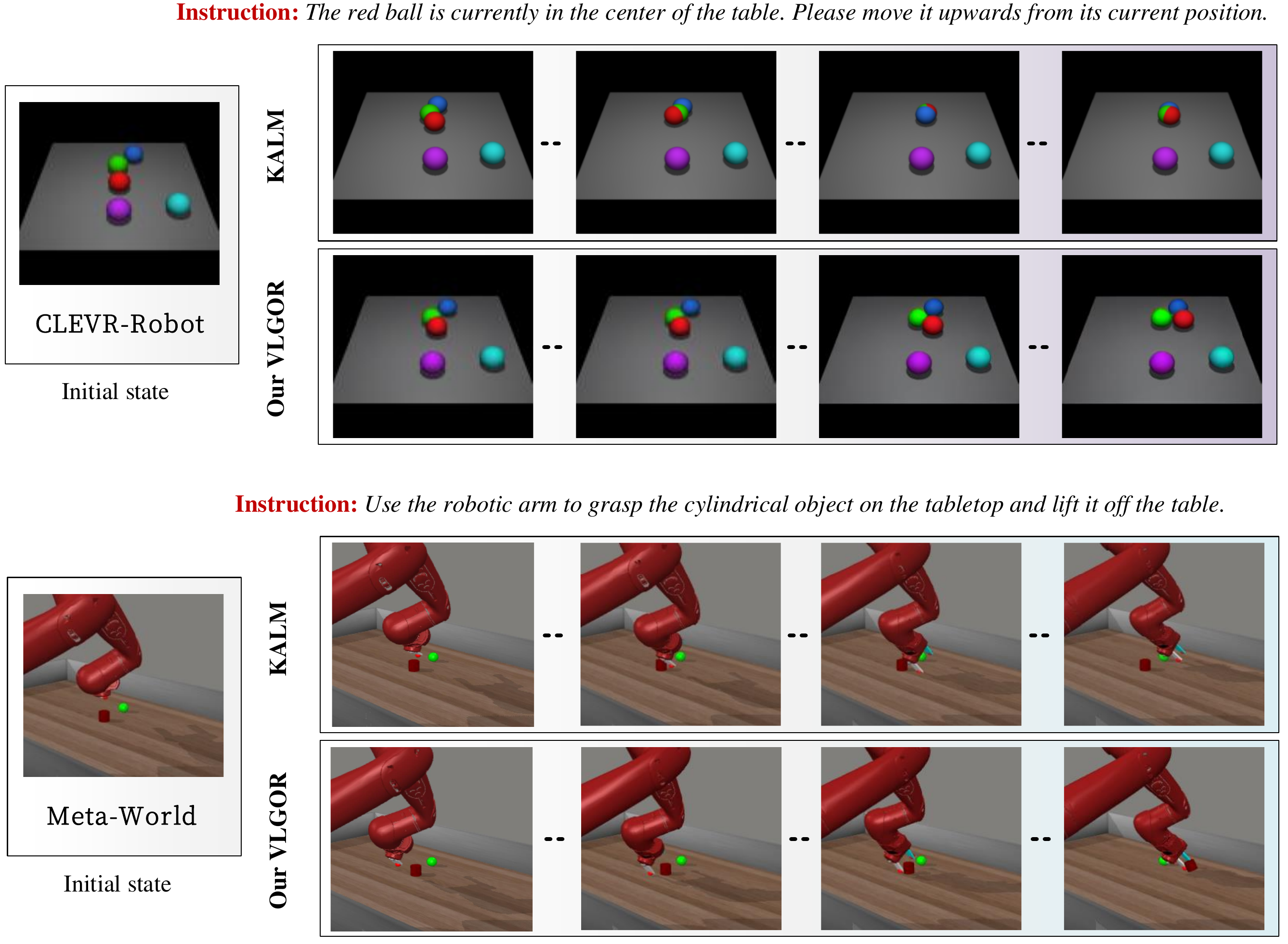}}
    \caption{
     Comparison of imaginary rollouts generated by KALM~\cite{pang2024kalm} and the proposed VLGOR. 
    }
    \label{fig5}
  \end{center}
\end{figure*}

\begin{table*}[t]
\centering
\caption{Effect of the number of samples used for VLM finetuning. PSNR, SSIM, and MED evaluate the quality of generated rollouts, while Success Rate measures instruction execution. Lower MED is better, while higher values are preferred for the other metrics.}
\label{tab1}
\begin{tabular}{c cccc cccc}
\toprule
\multirow{2}{*}{Number of samples} 
& \multicolumn{4}{c}{CLEVR-Robot} 
& \multicolumn{4}{c}{Meta-World} \\
\cmidrule(lr){2-5} \cmidrule(lr){6-9}
& PSNR $\uparrow$ & SSIM $\uparrow$ & MED $\downarrow$ & Success rate $\uparrow$ 
& PSNR $\uparrow$ & SSIM $\uparrow$ & MED $\downarrow$ & Success rate $\uparrow$ \\
\midrule
5000  & 19.63 dB & 0.7426 & 0.1762 & 22.53\% & 17.41 dB & 0.6941 & 0.2015 & 16.15\% \\
10000 & 22.33 dB & 0.7915 & 0.1261 & 28.63\% & 21.25 dB & 0.7315 & 0.1548 & 20.56\% \\
10000 & 23.75 dB & 0.8072 & 0.0933 & 32.95\% & 22.18 dB & 0.7558 & 0.1334 & 23.14\% \\
15000 & 23.94 dB & 0.8123 & 0.0884 & 33.71\% & 22.64 dB & 0.7612 & 0.1285 & 23.57\% \\
\bottomrule
\end{tabular}
\end{table*}

\subsection{Experimental Setting}
\textbf{Robotic Manipulation Environments.} We evaluate VLGOR on widely used benchmarks, including CLEVR-Robot~\cite{jiang2019language} and Meta-World~\cite{yu2020meta}. In Meta-World, the agent controls a Sawyer robotic arm to manipulate various objects, whereas in CLEVR-Robot, the agent (silverpoint) manipulates five movable balls to achieve target configurations. We adopt the offline dataset collected by KALM~\cite{pang2024kalm}, which comprises 100,000 rollout-goal pairs. Environment-provided reward functions are used to obtain and integrate rewards for both the offline dataset and the generated imaginary rollouts.

\begin{figure}[ht]
  \vskip 0.2in
  \begin{center}
\centerline{\includegraphics[width=0.98\linewidth]{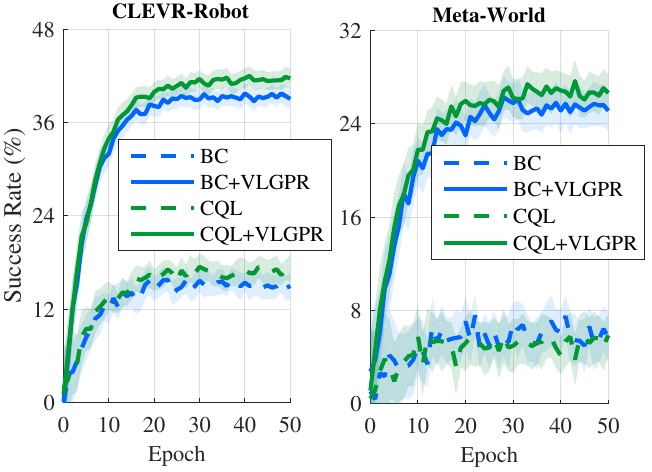}}
    \caption{
      Success rate curves on unseen counterfactual tasks. The \textit{x}-axis denotes training epochs, while the \textit{y}-axis denotes success rates of instruction completion in offline RL. Shaded regions represent half of the standard deviation over three random seeds.
    }
    \label{fig6}
  \end{center}
\end{figure}

\textbf{Definition of Unseen Tasks.} To evaluate the generalization capability of policies learned by VLGOR on novel tasks, we define three levels of evaluation tasks with increasing difficulty: (1) Instruction Rephrasing Tasks: The original natural language instructions in the offline dataset are paraphrased for the agent to perform the same manipulation tasks. (2) Counterfactual Tasks: Instructions are redefined via counterfactual prompting, modifying the target and/or required behavior to create manipulation tasks that are not present in the offline dataset. (3) Hard Tasks: The agent is required to solve tasks that differ substantially from those in the offline dataset, involving more complex combinations of target specifications and behavioral variations.

\textbf{Rollout Generation.} Following prior work~\cite{pang2023natural}, we use ChatGPT to generate natural language goals, resulting in 1,440 goals for CLEVR-Robot and 200 goals for Meta-World. During VLM fine-tuning, we construct a training set comprising 10,000 trajectories. For offline reinforcement learning, baseline methods are trained solely on the offline dataset with 6,400 rollout–goal pairs, whereas VLGOR additionally generates 5,400, 6,400, and 1,680 imaginary rollouts for the three levels of unseen tasks, respectively. When training policies for novel goals, VLGOR jointly leverages the offline dataset and the corresponding generated rollouts, maintaining a fixed ratio between real offline data and imaginary rollouts in each training batch.

\textbf{Implementation Details.} For environment grounding, we adopt BAGEL~\cite{deng2025emerging} as the VLM backbone and train it for 20 epochs with a batch size of 16. Offline RL baseline methods are implemented using d3rlpy~\cite{seno2022d3rlpy}, and policy optimization uses the Adam optimizer. All methods train policies for 50 epochs. Offline RL experiments are repeated with three different random seeds to ensure robustness. All experiments are conducted on a single NVIDIA H800 80GB PCIe GPU. Additional implementation details are provided in Appendix~\ref{appendixA}.

\subsection{Offline RL Performance Comparison}\label{sec42}

\cref{fig4} presents the success rates of different methods on unseen tasks of varying difficulty in the CLEVR-Robot and Meta-World environments. To ensure a fair comparison, the total data used for offline RL training is kept constant. Overall, methods augmented with rollout generation consistently outperform their corresponding offline RL baselines. Furthermore, the performance gains of VLGOR-enhanced methods over the baselines become increasingly pronounced as task difficulty rises.

\textbf{(1) For instruction rephrasing tasks}, both KALM~\cite{pang2024kalm} and VLGOR based methods achieve clear improvements over BC and CQL baselines in both environments. On CLEVR-Robot, the BC baseline attains a success rate of around 25\%, while BC+KALM and BC+VLGOR improve performance to approximately 41\%, with only a small gap between the two. Similar trends are observed for CQL-based methods. This is expected, as instruction rephrasing preserves the underlying task dynamics and intended goals, differing only in linguistic expression. In this context, language-based rollout generation effectively captures task-relevant dynamics, achieving performance comparable to or slightly exceeding that of VLGOR.

\textbf{(2) For counterfactual tasks}, VLGOR exhibits a clear advantage over both the baselines and KALM. On CLEVR-Robot, BC+VLGOR increases the success rate from approximately 15\% (BC) and 33\% (BC+KALM) to nearly 39\%. On Meta-World, VLGOR further outperforms KALM by roughly 10-14 percentage points. This improvement primarily stems from the physically and temporally consistent rollouts, as well as the effective integration of visual information into offline RL, which together facilitate more effective skill acquisition for complex and unseen tasks.

\textbf{(2) For hard tasks}, the benefits of VLGOR become most pronounced. On CLEVR-Robot, BC+VLGOR and CQL+VLGOR achieve success rates of approximately 19\% and 22\%, respectively, compared to 1–8\% for the baselines and KALM-based variants. Similar trends are observed on Meta-World. This improvement can be attributed to the use of physically and temporally consistent rollouts, as well as the integration of visual information into offline RL, which together facilitate more effective skill acquisition for complex and long-horizon tasks.

\subsection{Rollout Generation Quality Analysis}\label{sec43}
This section evaluates the quality of generated rollouts and their impact on offline RL through both qualitative visual comparisons and quantitative metrics. 

\begin{table*}[t]
\centering
\caption{Impact of different counterfactual prompts as instructions on task success rates (\%).}
\label{tab2}
\begin{tabular}{r ccc ccc}
\toprule
\multirow{2}{*}{Method} 
& \multicolumn{3}{c}{CLEVR-Robot} 
& \multicolumn{3}{c}{Meta-World} \\
\cmidrule(lr){2-4} \cmidrule(lr){5-7}
& Rephrasing & Counterfactual & Hard 
& Rephrasing & Counterfactual & Hard  \\
\midrule
Baseline & 25.3 & 15.2 & 1.2  & 13.6 & 5.9  & 3.8  \\
\hline
w/o target    & 40.6 & 36.3 & 14.3 & 21.8 & 22.4 & 13.1 \\
w/o behavior  & 39.8 & 35.6 & 13.1 & 20.3 & 20.9 & 11.8 \\
Our VLGOR     & 41.7 & 39.3 & 19.3 & 23.6 & 25.6 & 17.8 \\
\bottomrule
\end{tabular}
\end{table*}

\textbf{Qualitative Visualization of Generated Rollouts.} To evaluate the fidelity of generated rollouts, \cref{fig5} compares imaginary rollouts in CLEVR-Robot and Meta-World. KALM uses language to describe states for generating rollouts, while for comparison, each frame is rendered using the MuJoCo physics engine~\cite{todorov2012mujoco}. In CLEVR-Robot, the task is to move the red ball upward. KALM correctly identifies the red ball and moves it incrementally, but fails to fully capture spatial relationships and inter-object dynamics when the red ball interacts with the green and blue balls. These errors accumulate along the trajectory, eventually causing the three balls to overlap. In Meta-World, the task is for the robotic arm to grasp and lift a cylinder on the table. KALM fails to accurately localize the cylinder, producing trajectories that do not complete the task.

In contrast to KALM, the proposed VLGOR method explicitly models object spatial configurations and enforces temporal consistency when generating trajectories, yielding rollouts that are both physically plausible and visually coherent. These results indicate that the LLM possesses a nontrivial understanding of the environment and can reason effectively about object dynamics, thereby providing a foundation for acquiring more complex skills.

\textbf{Quantitative Evaluation of Rollout Quality.} We further assess the quality of generated rollouts and their impact on offline RL. During VLM finetuning, we leverage rollout-goal pairs from KALM. \cref{tab1} reports how the number of finetuning samples affects performance in CLEVR-Robot and Meta-World. Rollout quality is evaluated using PSNR and SSIM~\cite{wang2004image}, the Euclidean distance (MED) between the main object and ground truth (with positions normalized), and Success Rate for instruction execution.

From the results in~\cref{tab1}, it can be observed that all evaluation metrics consistently improve as the number of finetuning samples increases. In particular, PSNR and SSIM steadily rise, while MED decreases, indicating that more finetuning data substantially improves the quality of the generated rollouts. Meanwhile, the success rate of instruction execution also increases, suggesting that higher-quality rollouts contribute to more reliable policy execution. Notably, when the number of finetuning samples exceeds 10k, performance improvements become marginal, indicating that the benefits of adding more finetuning data gradually saturate.

\subsection{Model Analysis}\label{sec44}

\textbf{Training Curves of Different Methods.} We first present the success rate curves of different methods on unseen counterfactual tasks over training epochs. As shown in~\cref{fig6}, on CLEVR-Robot and Meta-World, all methods exhibit a rapid increase in success rates during the early training stage, followed by gradual convergence. In environments such as Meta-World, the high complexity of the tasks leads to relatively larger fluctuations in the success rate curves. Compared with baseline methods, those augmented with VLGPR maintain higher success rates throughout training and exhibit smaller performance variance, indicating improved stability and generalization in instruction completion.

\textbf{Impact of Different Counterfactual Instructions.} To assess the effect of different counterfactual prompts on task success, we conduct experiments in the CLEVR-Robot and Meta-World environments. As shown in~\cref{tab2}: (1) As the baseline, the BC algorithm, which uses only the original interaction data, performs worst across all tasks, highlighting that limited interaction data alone is insufficient for unseen complex scenarios. (2) Removing either the counterfactual target (w/o target) or behavior (w/o behavior) prompts improves success rates, with diverse counterfactual behaviors yielding the largest gains, indicating their role in generating more informative rollouts. (3) VLGOR achieves the highest success rates across all environments by jointly leveraging both target and behavior prompts, producing diverse rollouts and enhancing agent generalization to novel tasks.

\section{Conclusion and Limitation}
In this work, we leverage a vision-language model to provide richer supervisory signals for offline reinforcement learning, addressing low-level control challenges in unseen tasks. We propose VLGOR, a novel method that generates more realistic imaginary rollouts. By using counterfactual prompts as instructions, VLGOR enhances the diversity of rollouts, supplying richer visual and linguistic information to offline RL and enabling agents to handle novel tasks. Experimental results on two robotic manipulation benchmarks demonstrate the effectiveness of VLGOR.

This study has two main limitations. First, our experiments are confined to relatively simple robotic control simulations. Future work will aim to extend VLGOR to more complex control scenarios and to develop more effective environment grounding to provide agents with reliable visual and linguistic knowledge. Second, VLGOR jointly generates states and actions, which requires it to simultaneously learn environment modeling and imitation policies. A potential solution is to adopt a progressive strategy that uses predicted state information to infer actions, ensuring consistency between the two while reducing the model’s workload.

\section*{Impact Statement}
This paper presents work whose goal is to advance the field of Machine
Learning. There are many potential societal consequences of our work, none
which we feel must be specifically highlighted here.

% In the unusual situation where you want a paper to appear in the
% references without citing it in the main text, use \nocite
%\nocite{langley00}

\bibliography{example_paper}
\bibliographystyle{icml2026}

%%%%%%%%%%%%%%%%%%%%%%%%%%%%%%%%%%%%%%%%%%%%%%%%%%%%%%%%%%%%%%%%%%%%%%%%%%%%%%%
%%%%%%%%%%%%%%%%%%%%%%%%%%%%%%%%%%%%%%%%%%%%%%%%%%%%%%%%%%%%%%%%%%%%%%%%%%%%%%%
% APPENDIX
%%%%%%%%%%%%%%%%%%%%%%%%%%%%%%%%%%%%%%%%%%%%%%%%%%%%%%%%%%%%%%%%%%%%%%%%%%%%%%%
%%%%%%%%%%%%%%%%%%%%%%%%%%%%%%%%%%%%%%%%%%%%%%%%%%%%%%%%%%%%%%%%%%%%%%%%%%%%%%%
\newpage
\appendix
\onecolumn

\section{Implementation Details.}\label{appendixA}

\subsection{Network Architecture}

\textbf{VLM for Rollout Generation.} 
The VLM is built upon the BAGEL framework, which employs a mixture-of-transformers architecture. Language instructions and the initial state are encoded using the Qwen2Tokenizer, while the initial visual state is represented as $128{\times}128$ images and encoded via a Vision Transformer (ViT) encoder. For text token prediction, BAGEL leverages the established strengths of autoregressive language models, whereas for visual token prediction, it adopts the rectified flow method. 
Additionally, a latent diffusion decoder with a Variational Autoencoder (VAE) is used to predict images corresponding to future states along the trajectory. During fine-tuning, Low-Rank Adaptation (LoRA) layers are inserted into the Multi-Modal Self-Attention (MMSA) module, and only these parameters are updated to facilitate cross-modal information modeling. In addition, the parameters of the two prediction heads are also updated to better capture modality-specific features.

\textbf{Policy Network for Offline RL.}  
For offline RL, textual inputs are first encoded by BERT into 768-dimensional feature vectors, which are then processed by a two-layer Multilayer Perceptron (MLP) consisting of Linear ($768{\to}512$) with ReLU activation, followed by Linear ($512{\to}256$) and LayerNorm. 
Visual state inputs are processed by a lightweight convolutional encoder comprising three convolutional layers: Conv ($3{\to}32$, kernel size 8, stride 4), Conv ($32{\to}64$, kernel size 4, stride 2), and Conv ($64{\to}64$, kernel size 3, stride 1), each followed by a Rectified Linear Unit (ReLU) activation. 
The extracted features are then flattened and projected to a 256-dimensional representation via a linear layer and LayerNorm. 
Finally, a fusion encoder integrates the textual and visual features, producing a 512-dimensional joint representation that serves as input to the policy network for subsequent reinforcement learning.

\subsection{Hyper-parameter Setting}

\begin{table}[H]
\centering
%\small
%\renewcommand{\arraystretch}{1.12}
\caption{VLM Hyper-parameter Configuration.}
\label{tab:app_vlm_infer}
\begin{tabular}{lc}
\toprule
\textbf{Hyper-parameter} & \textbf{Setting} \\
\midrule
Diffusion timesteps & 70 \\
Text temperature & 0.3 \\
Classifier-free guidance interval & [0.0, 1.0] \\
Maximum thinking tokens & 1024 \\
VAE preprocess size & $256\times256$ \\
ViT preprocess size & $224\times224$ \\ 
Latent patch size & 2 \\
Max latent size & 64 \\
\bottomrule
\end{tabular}
\end{table}

\begin{table}[H]
\centering
%\small
%\renewcommand{\arraystretch}{1.12}
\caption{Offine RL Hyper-parameter Configuration.}
\label{tab:app_bc}
\begin{tabular}{lc}
\toprule
\textbf{Hyper-parameter} & \textbf{Setting} \\
\midrule
Number of training epochs & 50 \\
Batch size & 128 \\
Learning rate & $3\times 10^{-4}$ \\
Minimum learning rate & $1\times 10^{-6}$ \\
Learning rate schedule & Cosine annealing \\
Stream fit & Yes \\
Merge fit & No \\
\bottomrule
\end{tabular}
\end{table}

\section{Specification of Actions and Instructions in Rollouts.}\label{bbb}

\subsection{Explanation of Actions in Rollouts.}
In Meta-World, the agent controls a Sawyer robot to manipulate various objects, such as doors, drawers, and windows. The target configurations in the offline dataset include: reaching (reach), pushing (push), picking and placing (pick-place), pressing a button (button-press), unlocking a door (door-unlock), opening a door (door-open), opening a window (window-open), turning on a faucet (faucet-open), pushing a coffee cup (coffee-push), and pressing a coffee machine button (coffee-button-press). The action space is $\mathbb{R}^{4}$, representing the gripper’s motion and open/close control.

In CLEVR-Robot, the agent (SilverPoint) manipulates five movable balls. The target configurations in the offline dataset involve moving a specific ball relative to a target ball in a specified direction (forward, backward, left, or right). The action space is $\mathbb{R}^{40}$, where each action is represented as a one-hot vector indicating which specific ball is moved in which direction.

\subsection{Explanation of Counterfactual Instructions in Rollouts.}
To generate diverse goal-conditioned rollouts, we adopt counterfactual instructions. Each constructed instruction consists of three components: a \emph{background} description $t$, a manipulation \emph{target} $m$, and a \emph{behavior} pattern $b$. This factorization enables systematic goal edits at a fixed visual state $s_t$.

Specifically, given an original instruction, we generate two types of feasible counterfactual instructions $G'{=}\langle t,m,b\rangle$ by modifying only a single component: (i) counterfactual target and (ii) counterfactual behavior. In our experiments, the manipulation target $m$ corresponds to selecting the object of interest, while the behavior pattern $b$ specifies the intended motion. We render original instruction into natural language using a fixed template that keeps $t$ constant while substituting $m$ and/or $b$. Under this scheme, the counterfactual target changes the target object while keeping the behavior unchanged, whereas the counterfactual behavior changes the behavior while keeping the target object unchanged. An example of the two generated counterfactual instructions in the CLEVR-Robot environment is shown in~\cref{tabins}.

\begin{table}[H]
\centering
\small
\renewcommand{\arraystretch}{1.2}
\caption{Examples of generated counterfactual instructions in the CLEVR-Robot environment. Red and blue fonts indicate the target and behavior, respectively.}
\label{tabins}
\begin{tabular}{l l l}
\toprule
\textbf{Original Instruction} & \textbf{Type} & \textbf{Generated Counterfactual Instruction} \\
\midrule
\multirow{2}{*}{Move the green ball toward the right.} 
& Counterfactual Target & Move the \textcolor{red}{blue ball} toward the right. \\
& Counterfactual Behavior & Move the green ball toward \textcolor{blue}{the left}. \\
\bottomrule
\end{tabular}
\end{table}

In the Meta-World environment, the generation of counterfactual instructions differs from that in the CLEVR-Robot environment. Since Meta-World tasks often involve only a single object of interest, replacing the target object is not feasible. Therefore, we generate counterfactual instructions by modifying the final goal of the behavior instead. An example of the two generated counterfactual instructions in the Meta-World environment is shown in~\cref{tabinmeta}.

\begin{table}[H]
\centering
\small
\renewcommand{\arraystretch}{1.2}
\caption{Examples of generated counterfactual instructions in the Meta-World environment.}
\label{tabinmeta}
\begin{tabular}{l l l}
\toprule
\textbf{Original Instruction} & \textbf{Type} & \textbf{Generated Counterfactual Instruction} \\
\midrule
\multirow{2}{*}{Move the gripper to the target.} 
& Counterfactual Target & Move the gripper to \textcolor{red}{the left of the target position}. \\
& Counterfactual Behavior & Move the gripper to the target region \textcolor{blue}{along a straight-line trajectory}. \\
\bottomrule
\end{tabular}
\end{table}

\section{Experimental Results}
\subsection{Visualization of Rollouts Generated by Different Instructions.}
\begin{figure}[H]
    \centering
    \includegraphics[width=0.98\linewidth]{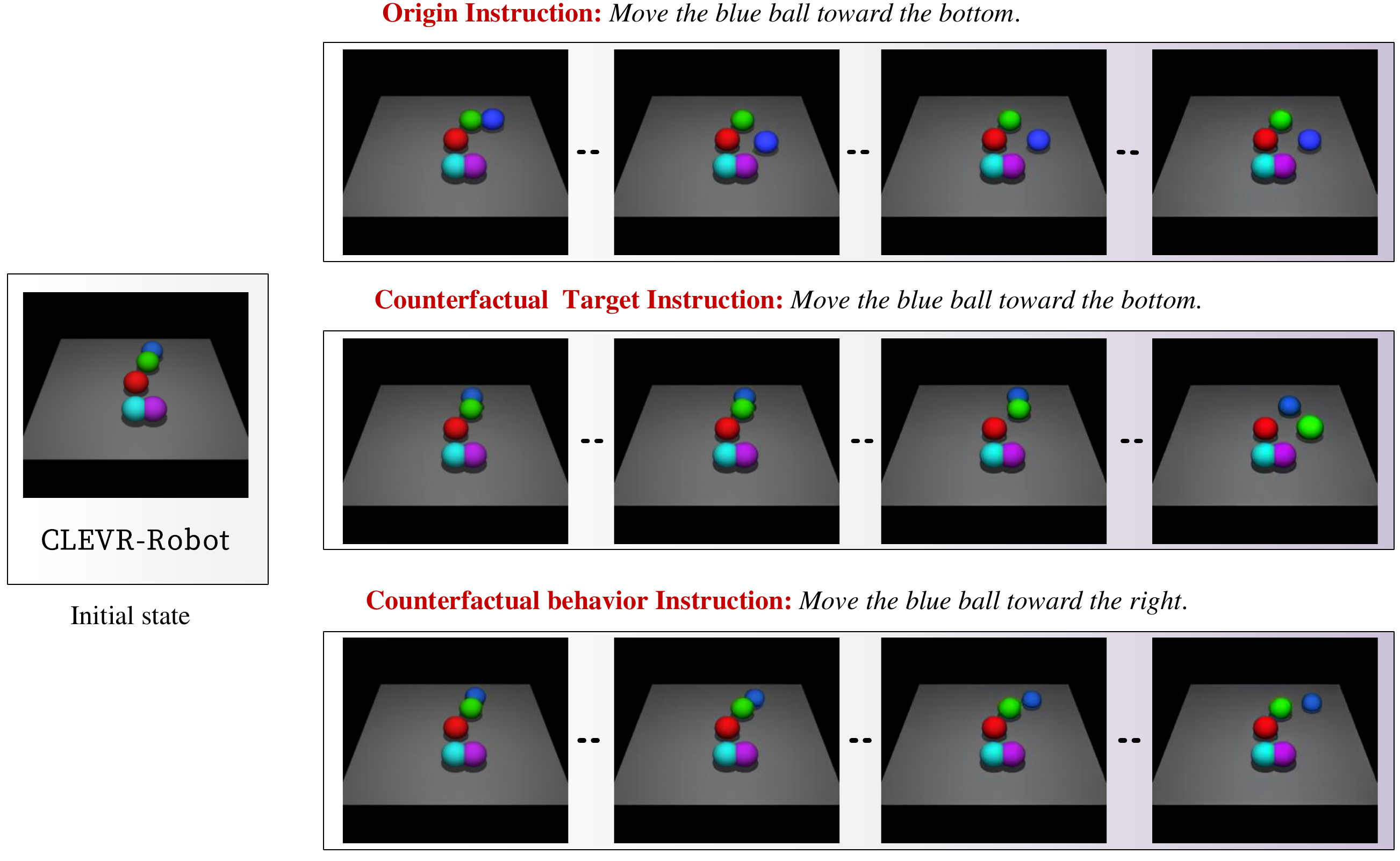}
    \caption{Visualization of rollouts generated by different instructions.}
    \label{figexample1}
\end{figure}

Given the same initial state, we generate rollouts using both counterfactual target and counterfactual behavior instructions.~\cref{figexample1} presents the visualizations of rollouts produced under these two types of instructions. Both counterfactual instruction types generate realistic rollouts while effectively increasing the diversity of interaction data.

\subsection{Experimental Evaluation of VLGOR with Different Offline RL Algorithms.}
\begin{figure}[H]
    \centering
    \includegraphics[width=0.88\linewidth]{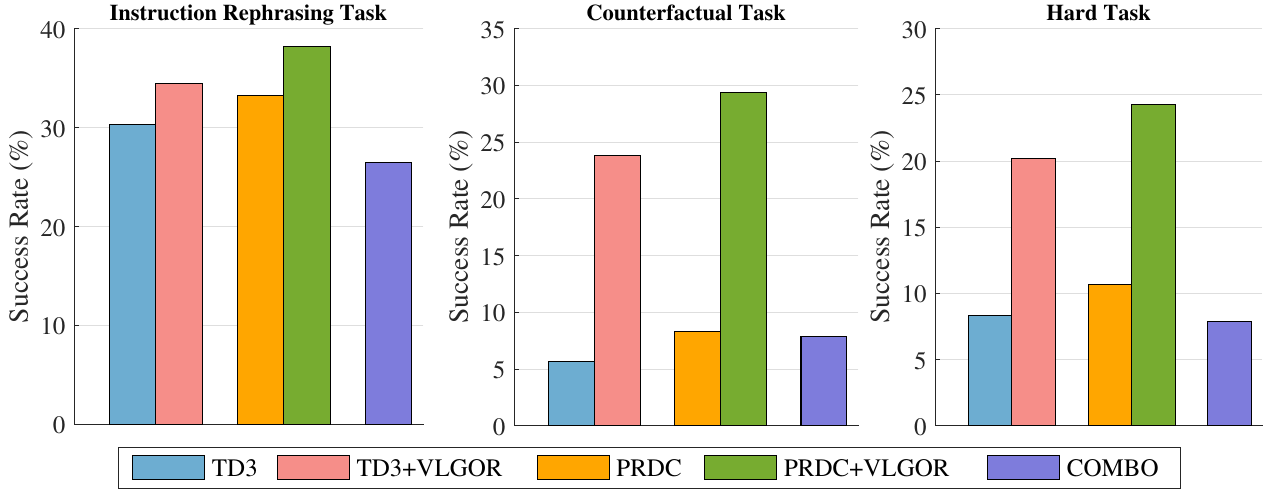}
    \caption{Success rates (\%) of different methods in the Meta-World environment.}
    \label{figexample2}
\end{figure}

\cref{figexample2} reports the success rates of different offline RL methods in the Meta-World environments. 
We evaluate VLGOR in combination with two offline RL algorithms: TD3~\cite{fujimoto2021minimalist}, which constrains the policy toward the demonstrated behaviors while leveraging the stability of TD3, and PRDC~\cite{ran2023policy}, which regularizes the policy toward nearby state–action pairs in the offline dataset via tree search. 
For comparison, we also include COMBO~\cite{yu2021combo}, a model-based offline RL method that employs an ensemble of environment models to achieve conservative policy learning. 
Overall, VLGOR consistently improves performance over the corresponding baselines and achieves higher success rates than COMBO on unseen tasks.

%%%%%%%%%%%%%%%%%%%%%%%%%%%%%%%%%%%%%%%%%%%%%%%%%%%%%%%%%%%%%%%%%%%%%%%%%%%%%%%
%%%%%%%%%%%%%%%%%%%%%%%%%%%%%%%%%%%%%%%%%%%%%%%%%%%%%%%%%%%%%%%%%%%%%%%%%%%%%%%

\end{document}